\documentclass{esannV2}
\usepackage{amssymb,amsmath,array}
\usepackage{graphicx}
\usepackage{booktabs}
\usepackage{siunitx}
\usepackage{float}

\begin{document}
\title{Towards Learning a Generalizable 3D Scene Representation from 2D Observations}

\author{Martin Gromniak$^{1,2}$, Jan-Gerrit Habekost$^1$, Sebastian Kamp$^1$, \\ Sven Magg$^3$ and Stefan Wermter$^1$
%
%
%
\vspace{.3cm}\\
%
1 - University of Hamburg - Department of Informatics, Hamburg - Germany
%
\vspace{.1cm}\\
2 - ZAL Center of Applied Aeronautical Research, Hamburg - Germany
\vspace{.1cm}\\
3 - Hamburger Informatik Technologie-Center e.V. (HITeC), Hamburg - Germany\\
}

\maketitle

\begin{abstract}
We introduce a Generalizable Neural Radiance Field approach for predicting 3D workspace occupancy from egocentric robot observations. Unlike prior methods operating in camera-centric coordinates, our model constructs occupancy representations in a global workspace frame, making it directly applicable to robotic manipulation. The model integrates flexible source views and generalizes to unseen object arrangements without scene-specific finetuning. We demonstrate the approach on a humanoid robot and evaluate predicted geometry against 3D sensor ground truth. Trained on 40 real scenes, our model achieves 26mm reconstruction error, including occluded regions, validating its ability to infer complete 3D occupancy beyond traditional stereo vision methods.
\end{abstract}

\section{Introduction}

Robotic manipulation requires understanding 3D scene geometry to plan grasps, predict collisions, and reason about object interactions.
However, contemporary computer vision pipelines operate primarily on 2D images using 2D backbones, even when target tasks demand explicit 3D reasoning. We advocate a 3D-first paradigm for embedded agents: (i) build 3D scene representations from egocentric observations in a self-supervised way, and (ii) use these representations for downstream perception and control. Making 2D-to-3D lifting explicit localizes geometric errors and provides a common world model serving multiple tasks.

Neural Radiance Fields (NeRFs) \cite{mildenhallNeRFRepresentingScenes2022} have emerged as a compelling solution for learning such 3D representations. Through novel view synthesis, NeRFs learn continuous density fields that are consistent with observed 2D images. Recent work has begun exploring NeRFs for robotics applications \cite{wangNeRFRoboticsSurvey2024}. Through movement, embodied agents can build 3D scene representations from egocentric visual observations without access to ground-truth geometry.

We present a novel Generalizable Neural Radiance Field approach for predicting 3D workspace occupancy from 2D views. Our method, trained on a collection of scenes, generalizes to new scenes without finetuning. Unlike traditional stereo methods, our approach infers complete geometry, including occluded regions, from sparse egocentric viewpoints. Our main contributions are: a generalizable, workspace-centric, and geometrically consistent 3D occupancy predictor, a successful implementation and demonstration on a humanoid robot platform, and a quantitative evaluation of the predicted 3D geometry using depth sensors.

\section{Background}

Neural Fields map coordinates to target signals. Initially proposed for 3D shape representation, they gained attention with Neural Radiance Fields \cite{mildenhallNeRFRepresentingScenes2022}, which map 3D positions and viewing directions to color and volume density. NeRFs are supervised by 2D images and camera poses, learning density fields that produce accurate volume renderings. However, standard NeRFs overfit to individual scenes and require retraining for new environments.

Generalizable NeRFs (GNeRFs) are trained across many scenes, enabling direct inference on unseen scenes without finetuning. Our work builds on MVSNeRF \cite{chenMVSNeRFFastGeneralizable2021} which constructs plane-swept cost volumes aggregating features across views in a single camera frustum. Different from this approach, we represent scenes in a workspace coordinate frame and evaluate the approach for geometrically accurate 3D reconstruction rather than novel view synthesis.

Previous work has explored occupancy prediction with NeRFs. Density Fields \cite{wimbauerScenesDensityFields2023} estimates 3D density from a single view but requires nearby views during training to retrieve color values. OccNeRF \cite{zhangOccNeRFAdvancing3D2024a} performs a 3D occupancy prediction for autonomous driving, leveraging photometric and semantic consistency between frames. In contrast, our approach is tailored to a robot workspace and learns from photometric consistency alone.

\section{Method}

\subsection{3D Occupancy Predictor}

We propose a 3D occupancy representation that can be estimated from one or multiple subsequent egocentric \textit{source} views of a scene. Our predictor consists of three neural modules trained end-to-end using image reconstruction loss between a \textit{target} view and its neural rendering. During training, we sample source and random target views; during inference, we use available source views. Figure 1 shows the module architecture.

\begin{figure}[t]
\centering
\includegraphics[width=1\textwidth]{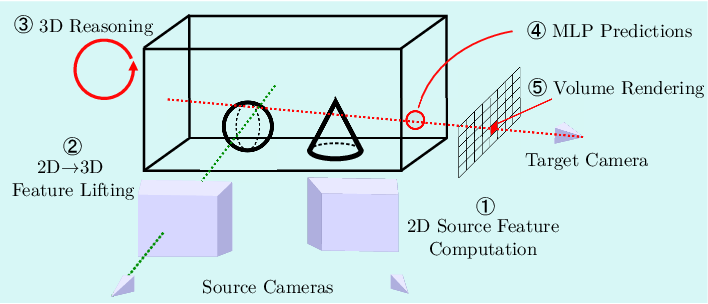}
\caption{Overview of the proposed neural architecture and data flow.}\label{Fig:MV}
\end{figure}

Following MVSNeRF \cite{chenMVSNeRFFastGeneralizable2021}, we employ three modules. First, a 2D feature extractor processes each source view using a lightweight 9-layer convolutional backbone with two downsampling stages, producing pixel-aligned 32-channel feature maps at H/4 $\times$ W/4 resolution. Second, we construct a cost volume in world coordinates with \SI{1}{\cm} voxel spacing. For each 3D point, we project it into each source view using the camera intrinsics and extrinsics and retrieve RGB-values and pixel-aligned features. We aggregate features across views by computing per-channel variance, yielding the cost volume voxel grid. Third, a 3D U-Net performs geometry reconstruction and occlusion reasoning. It progressively downsamples the input cost volume through three strided convolution stages, then upsamples back, outputting an 8-channel neural feature volume. Finally, a 6-channel, 128-width MLP uses bilinearly interpolated 3D volume features, relative positions within the volume, and view directions as inputs to regress density and RGB values, which are then used to calculate target pixel colors using standard volumetric rendering \cite{chenMVSNeRFFastGeneralizable2021}. During rendering, pixels with incomplete opacity accumulation receive background values from empty scene images, encouraging the model to allocate density only to changing scene elements. 
From the predicted densities, we compute opacity values at each point as $\alpha = 1 - \exp(-\sigma), \alpha \in [0,1]$ and treat those as our probabilistic occupancy.

\subsection{Humanoid Robot Setup and Dataset}

To demonstrate our approach, we use the humanoid robot NICOL (see Fig. 2). It has a 2-DoF 3D-printed head structure with two 4k RGB cameras and a RealSense D435if camera. Additionally, we employ three static RealSense cameras to capture the workspace from multiple perspectives. We performed intrinsic and extrinsic calibration of all cameras with respect to a world coordinate system that yields a reprojection error of $\sim2$ pixels.

We recorded 60 scenes, each containing a spatial arrangement of 5 objects from the YCB dataset \cite{calliBenchmarkingManipulationResearch}. Objects were repositioned between scenes. The robot scanned each workspace by moving its head through 15 positions from right to left, capturing 15 frames per head camera and one frame from each static workspace camera. The RealSense cameras record both RGB and depth data, and align both modalities. We use 20 scenes as a consistent evaluation set across all experiments. These scenes included occlusions where object parts were not visible from the head cameras.

\begin{figure}[t]
\centering
\includegraphics[width=1\textwidth]{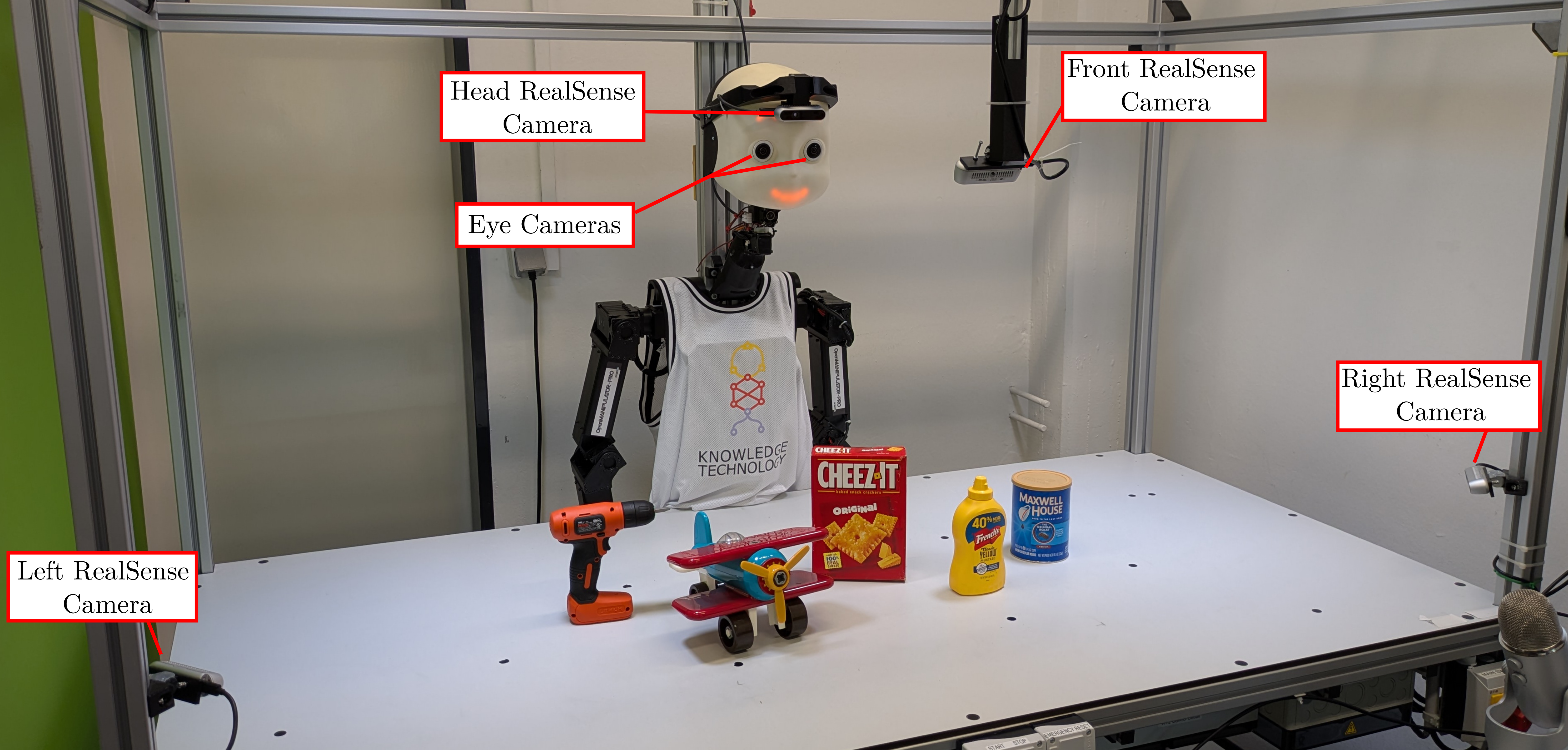}
\caption{NICOL robot and all involved workspace cameras. }\label{Fig:MV}
\end{figure}

\subsection{Training Setup}

During training, we randomly select one RealSense camera as the target view per step. If it is the head RealSense, we randomly sample one of 15 viewpoints. We render this target view and compute loss against the ground truth image. Opacity values within the workspace are predicted as a byproduct of rendering. Our loss combines Mean Squared Error on RGB values and the beta loss from Plenoxels \cite{fridovich-keilPlenoxelsRadianceFields2022}, which encourages binary alpha values: 
\vspace{\textheight}
\begin{equation}
L = L_{\text{MSE}} + 0.1 \times L_{\beta} = \frac{1}{N} \sum_{i=1}^{N} (I_i - \hat{I}_i)^2 + 0.1 \times (\alpha \times (1 - \alpha))
\end{equation}

We train the model for 400 epochs with learning rate of \(5 \times 10^{-4}\). 

\subsection{Evaluation Metrics}

We evaluate predicted 3D occupancy using the RealSense depth measurements as a proxy for full 3D ground truth. For each evaluation scene, we predict depth maps from four viewpoints: all workspace cameras and the center pose of the head RealSense camera. We compute predicted depth as $D = \min\{z_i : \sum_{j=0}^{i} \alpha_j \geq 0.5\}$ where $z_i$ are the sampled depths across one ray. We filter measurements to points within the workspace and create a validity mask. After applying this mask to both ground truth and predicted depth maps, we calculate the mean absolute error as a measure for 3D geometry reconstruction accuracy.

\section{Experiments}

We examine three source view settings: single view, stereo vision (one image from left and right eye camera), and three views from different head angles. Notably, stereo vision emerges as a two-view special case of our multi-view architecture. We also evaluate how performance scales with the number of training scenes $N_{train}$. We report the mean absolute depth error and the Peak signal-to-noise-ratio (PSNR), where the latter measures image rendering quality, in Table 1.
Three key observations can be made from our results. First, PSNR correlates strongly with geometric accuracy, with higher PSNR values corresponding to lower depth errors. Second, viewpoint diversity improves both metrics, indicating that our model effectively exploits additional visible geometry from varied camera poses. Third, depth error decreases as the number of training scenes increases, demonstrating that our approach benefits from greater scene diversity during training. Figures 3 and 4 show predictions for one evaluation scene.

\begin{table}[H]
\centering
\caption{Results for geometry prediction error and PSNR}
\label{tab:results}
\begin{tabular}{@{}lcccc@{}}
Source Cameras & Head Pose & $N_{train}$ & $MAE_{depth}$ & PSNR \\
\toprule
Left Eye & Mid & 40 & 0.03754 & 22.85 \\
Left + Right Eye & Mid & 40 & 0.03195 & 23.49 \\
Left Eye & Left+Mid+Right & 40 & \textbf{0.02638} & 23.64 \\
Left Eye & Left+Mid+Right & 20 & 0.03513 & 22.34 \\
Left Eye & Left+Mid+Right & 10 & 0.03800 & 21.48 \\
Left Eye & Left+Mid+Right & 5 & 0.04562 & 20.81 \\
\end{tabular}
\end{table}

\begin{figure}[H]
    \centering
    \includegraphics[width=1\linewidth]{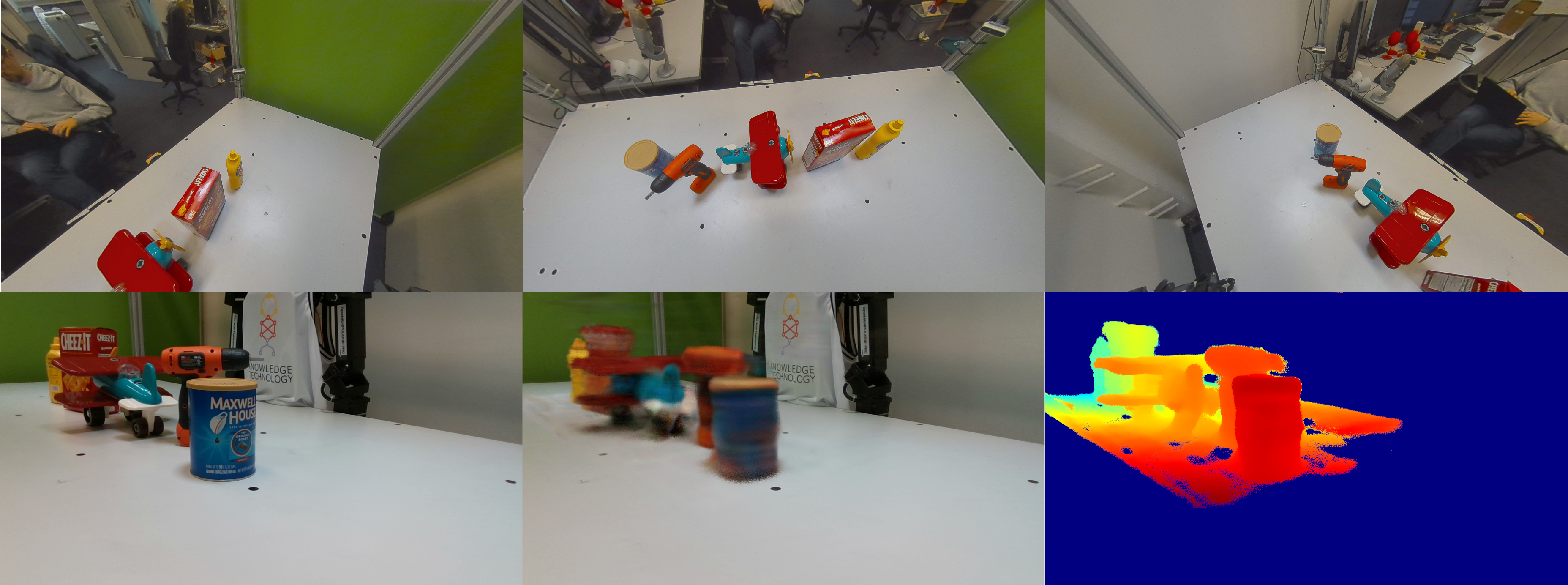}
    \caption{Top row: Three source views (Input to our model). Bottom left: Target view RealSense image. Bottom mid: Neural rendering for that view. Bottom right: Depth predictions by our model. Note that the geometry of the lower wing is correctly reconstructed despite not being  visible from the input views.}
\end{figure}  

\begin{figure}[H]
    \centering
    \includegraphics[width=1\linewidth]{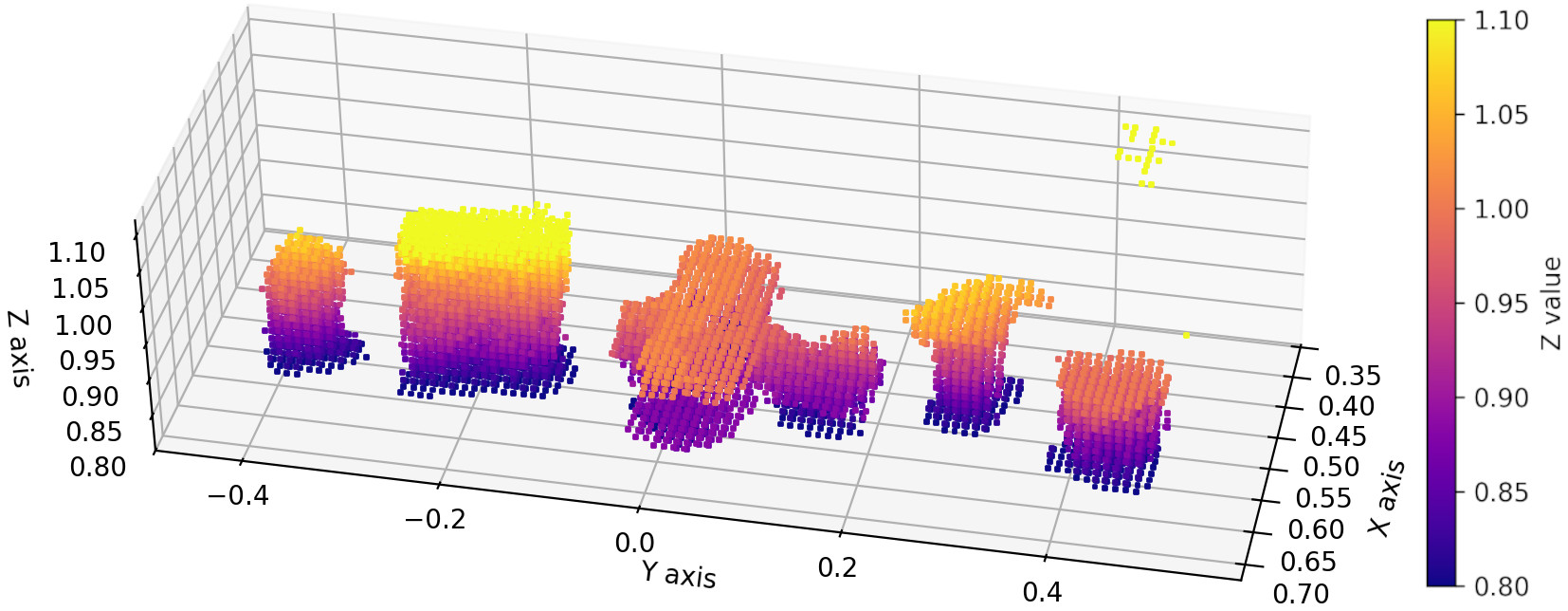}
    \caption{Prediction of the 3D occupancy for the same scene as in Figure 3.}
\end{figure}

\section{Conclusion}

We have introduced a flexible Generalizable NeRF model that learns geometrically accurate 3D workspace representations from 2D observations via novel view synthesis. Unlike prior GNeRF approaches that operate in camera-centric coordinate frames, our method constructs occupancy in a global workspace coordinate system, making it directly applicable to robotic manipulation tasks. The model flexibly integrates an arbitrary number of source views and demonstrates generalization to unseen object arrangements without scene-specific finetuning.

We plan to scale our dataset to include more object categories and scene complexity, examining knowledge transfer across object classes. The current single-channel occupancy representation provides a foundation for object manipulation tasks like grasping. We aim to extend the model to predict multi-channel semantic features, lifting 2D semantic information into the 3D representation using the same NeRF-based framework. This will enable task-relevant scene understanding directly in 3D space.

\bibliographystyle{unsrt}
\bibliography{bibliography} 

\end{document}